\documentclass[letterpaper, 10 pt, conference]{ieeeconf}  

\IEEEoverridecommandlockouts                              
\usepackage{epsfig}  
\usepackage{amsmath}
\usepackage{algorithm}
\usepackage{algorithmic}   
\usepackage{graphicx}
\usepackage{color}
\usepackage[dvipsnames]{xcolor}
\usepackage{amssymb}  
\usepackage[algo2e]{algorithm2e} 
\usepackage{tikz}
\usepackage{cite}
\usepackage{soul}
\usepackage{xcolor}
\usetikzlibrary{matrix,chains,positioning,decorations.pathreplacing,arrows}

\usepackage[pagebackref=true,colorlinks,allcolors=red]{hyperref}

\begin{document}
\title{\LARGE \bf
Deep Learning of Koopman Representation for Control
}

\author{Yiqiang Han$^{*}$, Wenjian Hao$^{*}$, and Umesh Vaidya
\thanks{$^{*}$These authors contributed equally to this work.} \thanks{All the authors are with the Department of Mechanical Engineering at Clemson University, Yiqiang Han is a research assistant Professor {\tt\small yigianh@g.clemson.edu}. Wenjian Hao is a graduate researcher, {\tt\small whao@g.clemson.edu},
Umesh Vaidya, is a professor of Mechanical Engineering, {\tt\small uvaidya@clemson.edu}. The first author wish to acknowledge research support from ONR award N00014-19-1-2295. The third author wish to acknowledge the technical and financial support of the Automotive Research Center (ARC) in accordance with Cooperative Agreement W56HZV-19-2-0001 U.S. Army CCDC Ground Vehicle Systems Center Warren, MI. }
}


\maketitle

\begin{abstract}

We develop a data-driven, model-free approach for the optimal control of the dynamical system. The proposed approach relies on the Deep Neural Network (DNN) based learning of Koopman operator for the purpose of control. In particular, DNN is employed for the data-driven identification of basis function used in the linear lifting of nonlinear control system dynamics. 
The controller synthesis is purely data-driven and does not rely on a priori domain knowledge. The OpenAI Gym environment, employed for Reinforcement Learning-based control design, is used for data generation and learning of Koopman operator in control setting. The method is applied to two classic dynamical systems on OpenAI Gym environment to demonstrate the capability.

\end{abstract}

\section{Introduction}

The problem of data-driven control of dynamic system is challenging with applications in robotics, manufacturing system, autonomous vehicles, and transportation networks. There is a shift from model-based to data-driven control paradigm with the increasing complexity of engineered systems and easy access to a large amount of sensor data. Deep reinforcement learning consisting of deploying deep neural networks for the learning of optimal control policy is emerging as one of the powerful tools for data-driven control of dynamic systems \cite{sutton2018reinforcement}. The controller synthesis for dynamic systems in the model-based and model-free setting has a long history in control theory literature. Furthermore, the problem of control design for a system with nonlinear dynamics is recognized to be a particularly challenging problem. More recently, the use of linear operators from the dynamical systems has offered some promises towards providing a systematic approach for nonlinear control
\cite{Meic_model_reduction, VaidyaMehtaTAC, Vaidya_CLM,raghunathan2014optimal,mezic_koopmanism,korda2018linear, susuki2011nonlinear,kaiser2017data,surana_observer,peitz2017koopman,mauroy2016global}. 

However, some challenges need to overcome for the successful extension of linear operator-based analysis methods to address the synthesis problem. The basic idea behind the linear operator framework involving Koopman  is to lift the finite-dimensional nonlinear dynamics to infinite-dimensional linear dynamics. The finite-dimensional approximation of the linear operator is obtained using the time series data of the dynamic system. 

One of the main challenges is the selection of appropriate choice of finite basis functions used in the finite but high dimensional linear lifting of nonlinear dynamics. The standard procedure is to make a-priori choice of basis function such as radial basis function or polynomial function for the finite approximation of the Koopman operator. However, such approach does not stand to take advantage of the known physics or the data in the choice of the basis function. An alternate approach based on the power of Deep Neural Network (DNN) could be used for the data-driven identification of the appropriate basis function in the autonomous (uncontrolled) setting \cite{lusch2018deep, yeung2019learning}. The conjecture is that the physical model can be retrieved from data observations  and DNN are efficient in exploiting the power of the data. In this paper, we extend the application of DNN for the selection of basis functions in a controlled dynamical system setting.

The main contributions of the paper are as follows. We extend the use of DNN for the finite dimensional representation of Koopman operator in controlled dynamical system setting. Our data-driven learning algorithm can automatically search for a rich set of suitable basis functions to construct the approximated linear model in the lifted space. The OpenAI Gym environment is employed for data generation and training of DNN to learn the basis functions \cite{brockman2016openai}. Example systems from OpenAI Gym environment employed for design of reinforcement learning control is used to demonstrate the capability of the developed framework. This work is on the frontier to bridge the understanding of model-based control and model-free control (e.g. Reinforcement Learning) from other domains.

\begin{figure*}[!htbp]
    \centering
    \includegraphics[width= 0.75\textwidth]{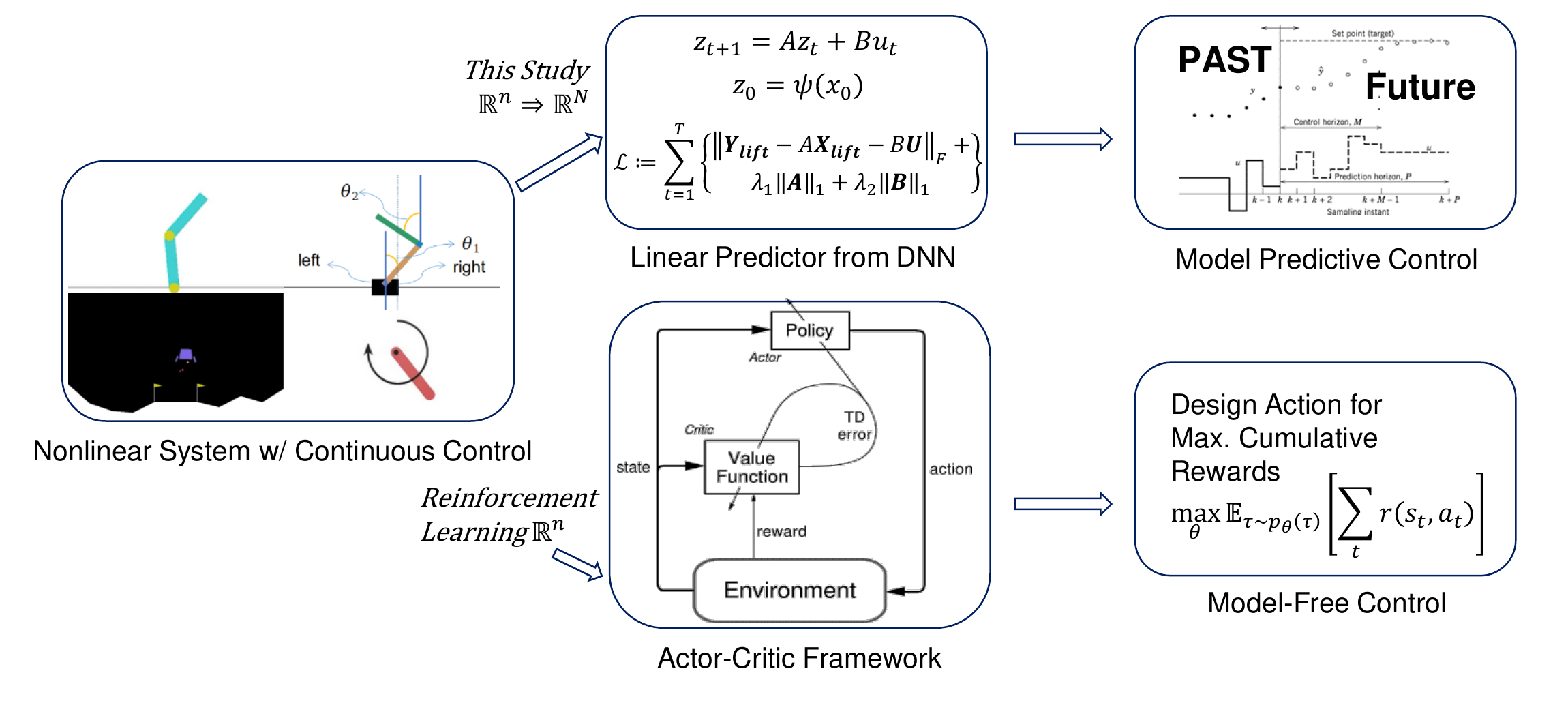}
    \caption{Deep Koopman Representation for Control (DKRC) framework with comparison to Reinforcement Learning}\label{fig:DKRC_scheme}
\end{figure*}

\section{Koopman Learning using Deep Neural Network}
Consider a discrete-time, nonlinear dynamical system along with a set of observation data of length $T$:
\begin{equation}\label{eqn.Koopman}
    \begin{gathered}
        x_{t+1}=F(x_t, u_t) \\
        {\scriptstyle\boldsymbol{X}=[x_1, x_2, \ldots , x_T], \quad  \boldsymbol{Y}=[y_1, y_2, ...  , y_T], \quad   \boldsymbol{U}=[u_1, u_2, \ldots , u_T] }
    \end{gathered}
\end{equation}  
where, $x_t\in \mathbb{R}^n$, $u_t\in \mathbb{R}^m$ is the state and control input respectively, and $y_t=x_{t+1}$ for $t=1,\ldots, K$. We use the capital letter $X$ to represent the collection of all data points in training set. The objective is to provide a linear lifting of the nonlinear dynamical system for the purpose of control. The Koopman operator ($\mathcal K$) is an infinite-dimensional linear operator defined on the space of functions as follows:

\begin{equation}\label{eqn.KoopmanOp}
[{\mathcal K}g](x)=g\circ { F}(x,u)
\end{equation}  
For more details on the theory of Koopman operator in dynamical system and control setting refer to \cite{koopman1931hamiltonian, williams2015data, williams2016extending, williams2014kernel}. Related work on extending Koopman operator methods to controlled dynamical systems can be found in References \cite{korda2018linear, proctor2016dynamic, kaiser2017data, kaiser2020data, broad2018learning, you2018deep,ma2019optimal}. While the Koopman based lifting for autonomous system is well understood and unique, the control dynamical system can be lifted in different ways. In particular, for control affine system the lifting in the functional space will be bi-linear \cite{huang2018feedback,ma2019optimal}. Given the computational focus of this paper we resort to linear lifting of control system. The linear lifting also has the advantage of using state of the art control methodologies such as linear quadratic regulator (LQR) and Model Predictive Control (MPC) from linear system theory. The linear lifting of nonlinear control system is given by

\begin{equation}\label{eqn.linearEqn}
    \Psi(x_{t+1})=A\Psi(x_t)+Bu_t
\end{equation}
where $\Psi(x)=[\psi_1(x),\ldots, \psi_N(x)]^\top \in \mathbb{R}^{N>>n}$ is the state or the basis in the lifted space. The matrix $A\in \mathbb{R}^{N\times N}$   and $B\in \mathbb{R}^{N\times m}$ are the finite dimensional approximation of the Koopman operator for control system in the lifted space. One of the main challenges in the linear lifting of the nonlinear dynamics is the proper choice of basis functions used for lifting. While the use of radial basis function, polynomials, and kernel functions is most common, the choice of appropriate basis function is still open. Lack of systematic approach for the selection of appropriate basis has motivated the use of Deep Neural Network (DNN) for data-driven learning of basis function for control.

The planned control inputs considered in this study are in continuous space and later will be compared to state-of-art model-free learning methods such as Reinforcement Learning. An overview of the scope of this work is shown in Figure \ref{fig:DKRC_scheme}. Both the two methods (our Deep Koopman Representation for Control (DKRC) approach and Reinforcement Learning) apply a data-driven model-free learning approach to learn the dynamical system. However, our approach seeks a linear representation of nonlinear dynamics and then design the control using a model-based approach, which will be discussed in later sections.

To enable deep Koopman representation for control, the first step is to use the DNN approach to approximate Koopman operator for control purposes. Our method seeks a multi-layer deep neural network to automate the process of finding the basis function for the linear lifting of nonlinear control dynamics via Koopman operator theory. Each feed-forward path within this neural network can be regarded as a nonlinear basis function ($\psi_i:\mathbb{R}^{n} \rightarrow \mathbb{R}$), whereas the DNN as a collection of basis functions can be written as $\Psi:\mathbb{R}^{n} \rightarrow \mathbb{R}^{N} $. Each output neuron is considered as a compounded result from all input neurons. Therefore, the mapping ensures that our Koopman operator maps functions of state space to functions of state space, not states to states. 

Unlike existing methods used for the approximation of the Koopman operator, where the basis functions are assumed to be known apriori, the optimization problem used in the approximation of Koopman operator using DNN is non-convex. The non-convexity arises due to simultaneous identification of the basis functions and the linear system matrices. 
 
During the training of DNN, the observables in state-space ($x$) are segmented into data pairs to guide the automated training process, whereas the output from the DNN is the representation of the lifting basis function for all data pairs ($\psi(x)$). This method is completely data-driven, and basis functions are learned directly from training datasets. 


The input dataset is split into three sets, whereas 70\% of the data are used as training samples, 15\% as validation, and another 15\% as testing cases. 
For the problems in this study, we choose a Multilayer Perceptron (MLP) module with four hidden layers to be used in between the neural network's input and output layers. At each layer in between the input and output layers, we apply hyperbolic-tangent (tanh) non-linear activation functions with biases at each neuron. 
In this study, all training processes are done using ADAM optimizer \cite{kingma2014adam} for neural networks parameters' backpropagation based on Pytorch 1.4 platform with NVIDIA V100 GPUs on an NVIDIA DGX-2 GPU supercomputer cluster. The deployment of the code has been tested on personal computers with NVIDIA GTX 10-series GPU cards.

\begin{figure}[!htbp]
    \centering
    \includegraphics[width=0.37\textwidth]{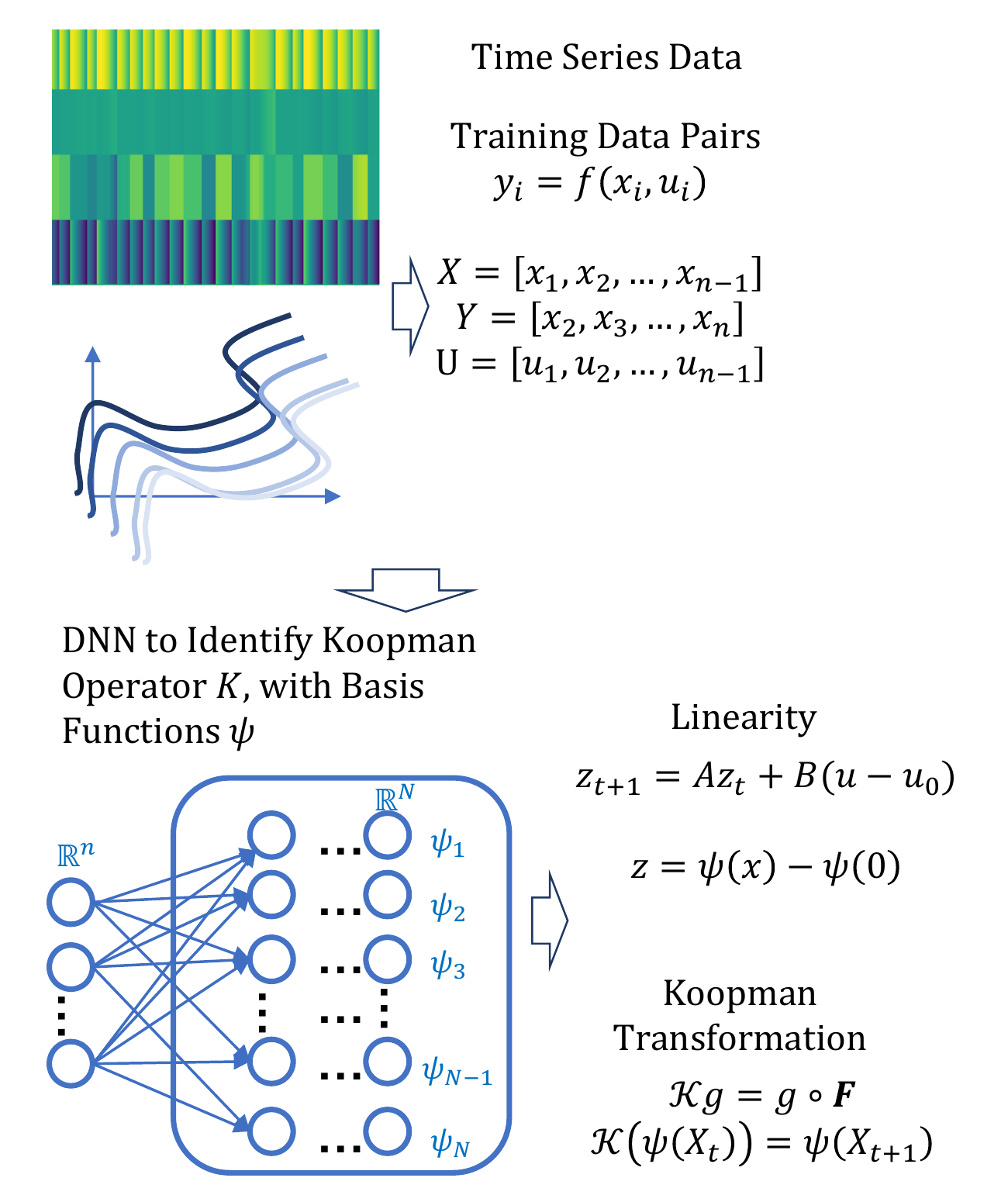}
    \caption{Koopman Operator Learning using Neural Network Scheme}\label{fig:network_scheme}
\end{figure}

During training, the DNN will iterate through epochs to minimize a loss term, where differences between two elements in lifted data pairs will be recorded and minimized in batch. At the convergence, we obtain a DNN, which automatically finds optimal lifting functions to ensure both the Koopman transformation (Equation \ref{eqn.Koopman}) and linearity (Equation \ref{eqn.linearEqn}) by tuning the DNN's loss function to a minimum. A schematic illustration of the Deep Koopman Representation learning can be found in Figure \ref{fig:network_scheme}.

The mapping from state-space to higher-dimensional space requires a modification to the existing control design governing equation, which is equivalent to regulation around a non-zero fixed point for non-linear systems. The affine transformation can be introduced without changing $A$ and $B$ matrices, as shown in Equation \ref{eqn.SystemEqn} below:
\begin{eqnarray}\label{eqn.SystemEqn}
    \begin{gathered}
        \boldsymbol z:=\Psi({\boldsymbol x})-\Psi_0,\;\;\;
        \boldsymbol v:=\boldsymbol u-u_0
    \end{gathered}
\end{eqnarray}
Here, we use $\Psi_0$ to represent this fixed point (in this case, goal position x=0) in the higher-dimensional space. To approximate $u_0$, it requires one additional auxiliary network for this constant. The network hyperparameters will then be trained together with the DNN.

After applying these changes, the Equation \ref{eqn.linearEqn} becomes the final expression of the high-dimensional linear model of the system in lifted space: 

\begin{equation}\label{eqn.zEqn}
    {\bf z}_{t+1}=A {\bf z}_t+ B{\bf v}_t
\end{equation}


Correspondingly, the problem we are solving becomes:
\begin{equation}\label{eqn.minCost}
    \min_{A,B,u_0} \sum_k \parallel {\bf z}_{t+1}-A{\bf z}_t-B{\bf (u)-u_0}\parallel_F
\end{equation}

The problem we are dealing with is non-convex. There is no one-shot solution to solve for multiple unknowns in the problem. We propose the following way to solve for the lifting function, with assumed $A$ and $B$ values at the beginning. We start the training with randomly initialized $A \in \mathbb{R}^{N \times N} $ and $B \in \mathbb{R}^{N \times m}$. The $\psi$ functions are learned within the DNN training, whereas the matrices $A$ and $B$ are kept frozen during each training epoch.

In particular, the following loss function, ${\cal L}$ in Equation \ref{eqn.DNN_loss}, is used in the training of DNN. 
\begin{eqnarray}\label{eqn.DNN_loss}
    {\cal L}:=&\sum\limits_{t=1}^T \parallel {\bf z}_{t+1}-A {\bf z}_{t}-B{\bf (u-u_0)} \parallel_F
\end{eqnarray}


At the end of each epoch, the $A$ and $B$ can be updated analytically by adding new information (with a learning rate of 0.5) based on the Equation \ref{eqn.findAB} as follows:
\begin{eqnarray}\label{eqn.findAB}
    [A,B]={\bf z}_{t+1}\begin{bmatrix}
        {\bf z}_{t} \\
           U 
         \end{bmatrix}\begin{pmatrix}\begin{bmatrix}
           {\bf z}_{t}\  
           U 
         \end{bmatrix}\begin{bmatrix}
           {\bf z}_{t} \\
           U 
         \end{bmatrix}\end{pmatrix}^\dagger
\end{eqnarray}
where $\dagger$ denotes pseudo-inverse.
In addition to finding system $A$ and $B$ matrix, following optimization problem is solved to compute the $C$ matrix mapping states from lifted space back to state space.
\begin{eqnarray}\label{eqn.C_matrix}
       & \min\limits_{C} \sum\limits_t \parallel x_t-C\Psi(x_t)\parallel_F ,\;\;\;
         &{\rm s.t.}\;\;\;C\Psi_0=0
\end{eqnarray}

\section{Koopman-based Control}

Based the insights gained from the linearized system in the higher-dimensional space, we can proceed to solve discrete LQR problem for the linear system with a quadratic cost:
\begin{eqnarray}\label{eqn.z_eqn}
    \begin{gathered}
        J(V) =\sum_t z_t^\top C^\top Q C z_t+v_t^\top R v_t
    \end{gathered}
\end{eqnarray}

Model Predictive Control (MPC) can also be implemented to accommodate problem constrains. Given a time horizon of $L$, we can solve for $V\in \mathbb{R}^{m\times L}$ by minimizing the following cost function, as illustrated in Equation \ref{eqn.MPCcost}. 
\begin{equation}\label{eqn.MPCcost}
\begin{gathered}
    J(V)= \sum_{t=0}^{L-1}(z_t^\top C^\top Q C z_t+v_t^\top R v_t) + z_L^\top C^\top Q C z_L \\
    \textit{ subject to} \begin{cases}
    v_t\leq u_{max}-u_0 \\ v_t\geq u_min-u_0
    \end{cases}
\end{gathered}
\end{equation}



To summarize, the algorithm of Deep Koopman Representation for Control (DKRC) is listed in Algorithm 1, where $L1(\theta)$ ensures a high dimension linear system, $L2(\theta)$ handles the controllability of the system.

To interprete the control strategy generated by DKRC, we proceed with the spectral analysis of the finite dimensional approximation. In particular, the eigenfunctions of the Koopman operator are approximated using the identified $A$ matrix. 

\begin{figure}[!htbp]
\centering
\includegraphics[width=0.3\textwidth]{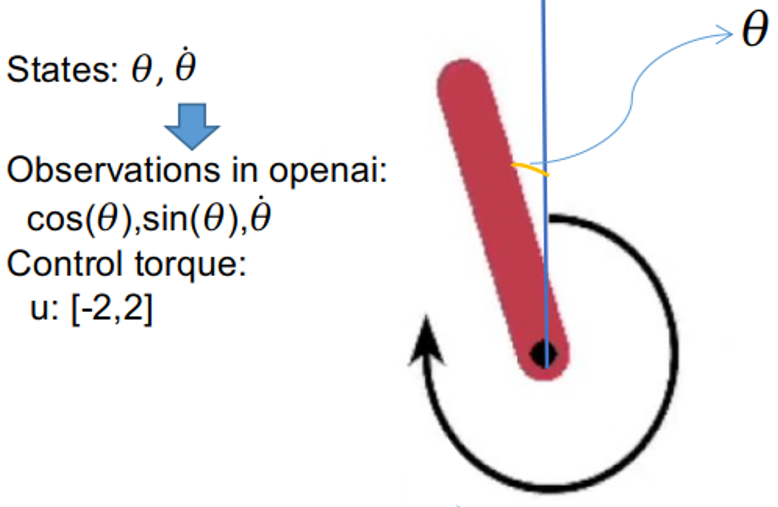}
\caption{Inverted Pendulum}\label{fig:pendulum_intro}
\end{figure}

The eigenfunctions learned from the neural network can be viewed as intrinsic coordinates for the lifted space and provide insights into the transformed dynamics in the higher dimensional space.
We use a classic inverted pendulum problem to demonstrate in this section. The pendulum system setup can be found in Figure \ref{fig:pendulum_intro}. We introduce a 3-D contour graph setting that uses the original state space observables (angular position \& angular velocity of the pendulum) as the basis and plot the eigenfunction values in the third dimension




\DontPrintSemicolon
\begin{algorithm}[H]
\SetAlgoLined
\KwIn{observations: x, control: u}
\KwOut{Planned trajectory and optimal control inputs: ($z_{plan}$, $v_{plan}$)}
\begin{itemize}
    \item Initialization 
        \begin{enumerate}
            \item Set goal position $x^*$
            \item Build Neural Network:
            $\psi_N(x_t;\theta)$
            \item Set $z(x_t;\theta) = \psi_N(x_t;\theta) - \psi_N(x^*;\theta)$
        \end{enumerate}
    \item Steps
    \begin{enumerate}
        \item Set $K=z(x_{t+1};\theta)*z(x_t;\theta)^\dagger$
        \item Set the first loss function $L1$\\
        $L_1(\theta)=\frac{1}{L-1}\sum_{t=0}^{L-1}\parallel z(x_{t+1};\theta)-K*z(x_t;\theta)\parallel$
        \item Set the second loss function $L_2$\\
        $[A,B]={\bf z}_{t+1}\begin{bmatrix}
            {\bf z}_{t} \\
               U 
             \end{bmatrix}\begin{pmatrix}\begin{bmatrix}
               {\bf z}_{t}\  
               U 
             \end{bmatrix}\begin{bmatrix}
               {\bf z}_{t} \\
               U 
             \end{bmatrix} \end{pmatrix}^\dagger $\\
        
         $L_2(\theta) = (N-rank(controllability(A,B)))+||A||_1+||B||_1$
         \item Train the neural network, updating the complete loss function\\ $L(\theta)=L_1(\theta)+L_2(\theta)$
         \item After converging, We can get system identity matrices A, B, C\\
         $C=\boldsymbol X_t*\psi_N(\boldsymbol X_t)^\dagger$
         \item Apply LQR or MPC control with constraints
    \end{enumerate}
\end{itemize}
\caption{Deep Koopman Representation for Control (DKRC)}
\end{algorithm}

\begin{figure}[!htbp]
\centering
\includegraphics[width=0.45\textwidth]{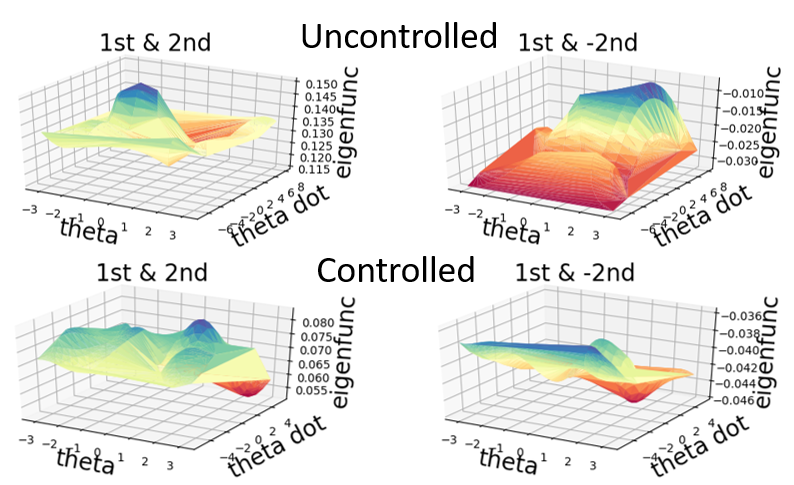}
\caption{Eigenfunctions learned from controlled inverted pendulum case. Left: real part of eigenfunctions,  Right: imaginary part of the eigenfunctions. The dominant eigenfunction has zero value imaginary part. }\label{fig:pendulum_eigenfunc_1n2only}
\end{figure}

An example of 3d plots for eigenfunctions obtained from DNN can be seen in Figure \ref{fig:pendulum_eigenfunc_1n2only}. The two images from the first row represent the first dominant eigenfunctions for uncontrolled pendulum case, whereas the second row represents the controlled case.

As can be seen, the first dominant eigenfunctions from both the two cases exhibit a very similar double-pole structure. The uncovered intrinsic coordinates exhibit a trend that is related to the energy levels within the system as base physical coordinates change, i.e., $\Dot{\theta}$ is positively related to the kinetic energy, whereas the $\theta$ is proportional to the potential energy. Towards the center of the 3d plot, both kinetic energy and potential energy exhibit declined absolute magnitudes. In the controlled case, torque at the joint is being added into the system's total energy count, which seeks a path to bring the energy of the system to the maximum in potential energy and lowest absolute value in kinetic energy. Due to the kinetic energy part ($\frac{1}{2}{\dot{\theta}}^2$) is dominating the numerical value of the energy, this effect has been reflected in the distribution of eigenfunction along the $\dot{\theta}$ axis. This comparison figure shows the feasibility of using eigenfunctions from learned Koopman operator to gain insight of the dynamical system, especially for the control purpose.

\section{Simulation Results}
We demonstrate the proposed continuous control method based on deep learning and Koopman operator (DKRC) on several example systems offered in OpenAI Gym\cite{brockman2016openai}. 
In this paper, we deploy model predictive control(MPC) on Inverted Pendulum, and adopt LQR for Lunar Lander (Continuous Control).

\subsection{Testing Environment}
The OpenAI Gym utilizes a Box2D physics engine, which provides a dynamical system for the testing. Gravity, object interaction, and physical contact behavior can be specified in the world configuration file. The dynamics of the simulation environment are unknown to the model. The model needs to learn the dynamics through data collected during training. Each simulation game is randomly initialized to generate initial disturbance.

\subsection{Reinforcement Learning}

The OpenAI Gym provides open-source environments primarily for Reinforcement Learning (RL) algorithm development purposes. Although RL shares the same model-free learning basis with our proposed DKRC method, the training processes are different. DKRC requires segmentation of data pairs from one batch of training data, whereas the RL will require a time series of data streams for model learning. Therefore it is hard to compare the learning process of the two algorithms side-by-side directly. In this study,  we are using a state-of-art Reinforcement Learning algorithm called Deep Deterministic Policy Gradient (DDPG) \cite{lillicrap2015continuous} to compare with results obtained using DKRC.  
The detailed comparison of the DKRC with controller designed using pre-determined choice of basis function such as using Extended Dynamic Mode Decomposition (EDMD) or DMD will be a subject of our future investigation.


\subsection{Inverted Pendulum}
The inverted pendulum environment is a classic 2-dimensional state space environment with one control input, as shown in Equation \ref{eqn.state_space_pen}. A visualization of the simulated environment is also shown in Figure \ref{fig:pendulum_intro}. The game starts with the pendulum pointing down with a randomized initial disturbance. The goal is to invert the pendulum to its upright position. In order to finish the game, the model needs to apply control input to drive the pendulum from start position ($\theta\in(-\pi,0)\cup(0,\pi)$, $\dot{\theta}=0$) to goal position ($\theta=0$, $\dot{\theta}=0$).


\begin{equation}\label{eqn.state_space_pen}
    \begin{gathered}
        \boldsymbol{\chi}=[\cos{\theta}, \sin{\theta}, \dot{\theta}], \quad where \  \theta\in[-\pi,\pi], \dot{\theta}\in[-8,8] \\
        \boldsymbol{U}=[u], \quad where \ u\in[-2,2]
    \end{gathered}
\end{equation}

In this case, the recorded $\theta$ orientation angle is in the form of a pair of cosine and sine values to ensure a smooth transition at $-\pi$ and $\pi$. The dynamical governing equation can be shown as follows:
\begin{equation}\label{eqn.pendulumGovEqn}
    ml^2\Ddot{\theta}=-\gamma\dot{\theta}+mglsin(\theta)+u
\end{equation}
This governing equation is unknown to our model, and the model needs to recover the physics-based on limited time-series data only. Moreover, the angular velocity ($\dot{\theta}$) and the torque input ($u$) at the shaft are also limited to a small range to add difficulties to finish the game. In most of the testing cases, we find that the pendulum needs to learn a swing back-and-forth motion before being able to collect enough momentum to swing up to the final vertical upright position. Although optimal control is possible, the trained model needs to adapt to the limitation imposed by the testing environment and make decisions solely based on state observations.

\begin{figure}[h]
\centering
\includegraphics[width=0.5\textwidth]{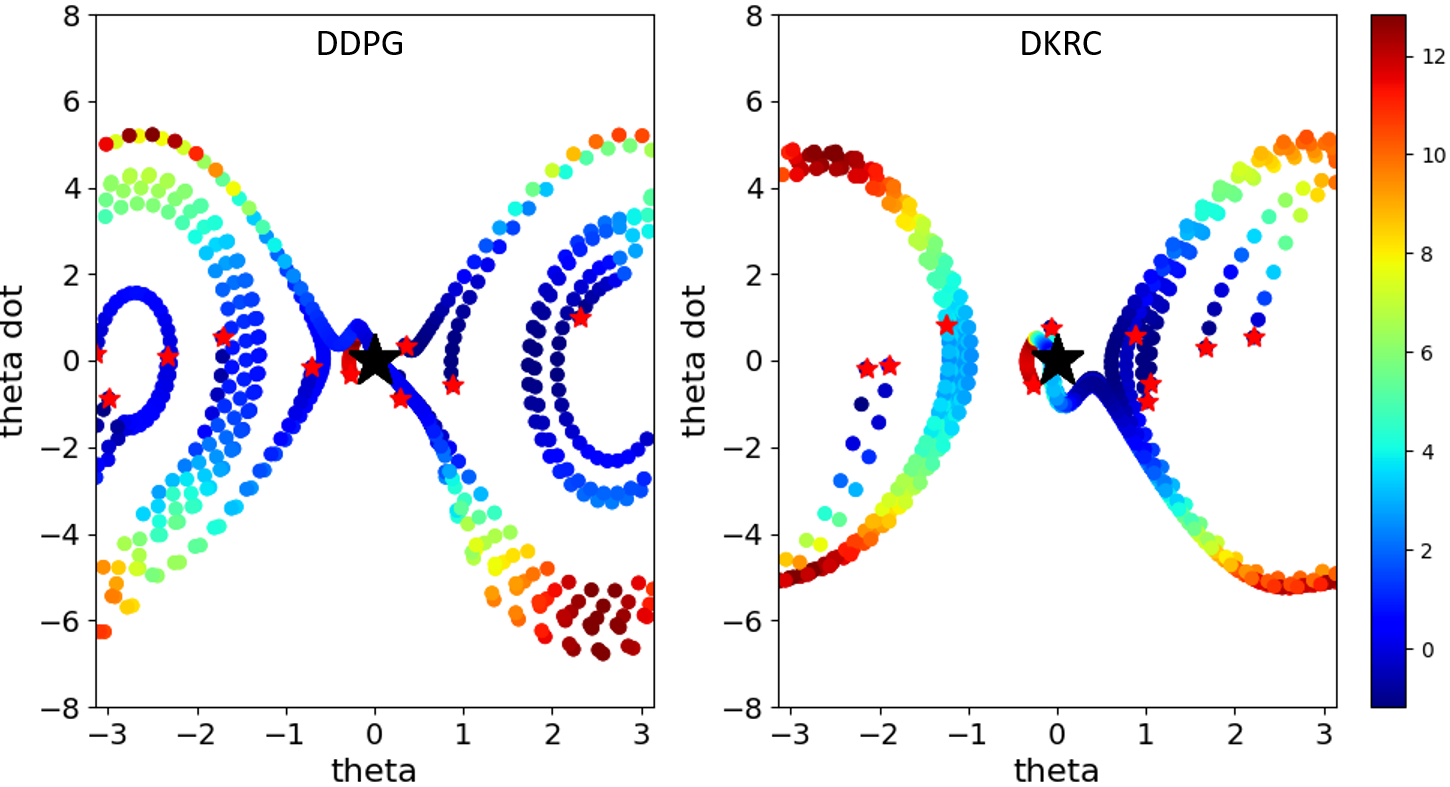}

\caption{Trajectories of swinging pendulum under control using DDPG reinforcement learning model (left) and the proposed DKRC algorithm (right). The figure exhibits the trajectories of the tip of the pendulum breaking through Hamiltonian energy level lines to arrive at the goal position ($\theta = 0$, $\dot{\theta}=0$)}\label{fig:pendulum_2dTraj_RL}
\end{figure}
\vspace{-10pt}
We train the DDPG reinforcement learning model first for benchmark purposes. Figure \ref{fig:pendulum_2dTraj_RL} shows the trajectory of the pendulum in 2D ($\theta$, $\dot{\theta}$) space based on trained DDPG algorithm. The black star in the center of Figure \ref{fig:pendulum_2dTraj_RL} is the goal position, and the concentric lines on both sides correspond to Hamiltonian total energy levels.  It is clear that due to the implicit model learned during the reinforcement learning process, the RL method cannot finish the game within a short time and therefore left many failed trails. 





\begin{figure}[h]
\centering
\includegraphics[width=0.45\textwidth]{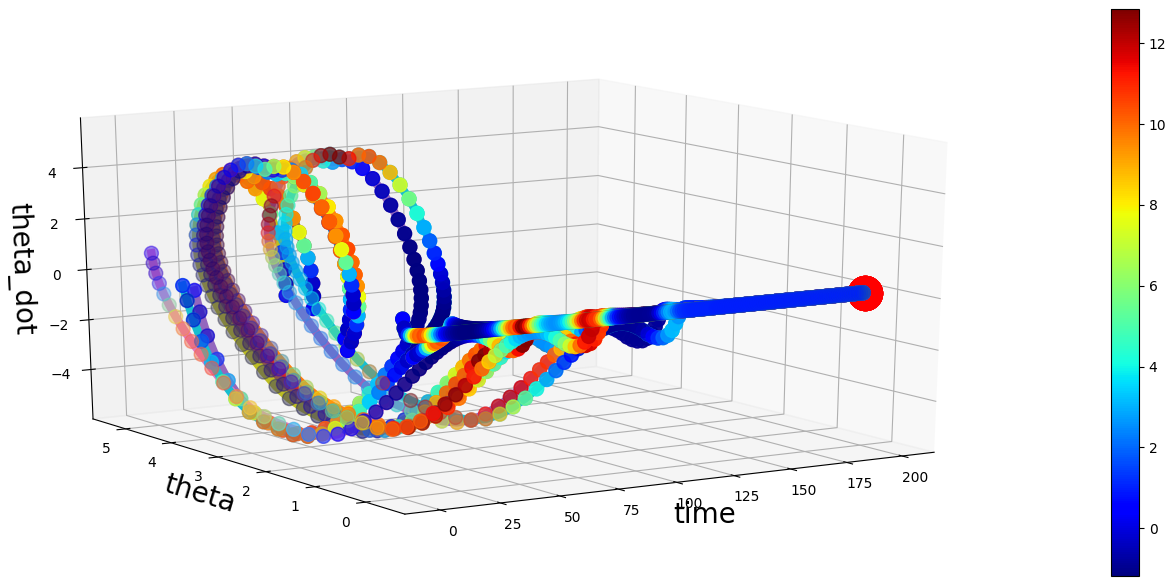}
\caption{Example 3D trajectories from 10 games played by DKRC, each game is reset by random initialization, 3D plot color-mapped by Hamiltonian energy level. Red Point: Goal states}\label{fig:pendulum_3dTraj}
\end{figure}

Besides the 2D comparison shown in Figure \ref{fig:pendulum_2dTraj_RL}, we are also presenting the state vs. time plots for the games played by DKRC model in 3D plot, as shown in Figure \ref{fig:pendulum_3dTraj}. In each deployment, the game starts with different initial conditions. The 3D plot is color-mapped by Hamiltonian energy levels. As indicated by the figure, the control generated by DKRC effectively minimized the system energy level within a short amount of deployment time. 

\vspace{-10pt}
\vspace{-10pt}

\subsection{Lunar Lander}
\begin{figure}[!htbp]
    \centering
    \includegraphics[width=0.4\textwidth]{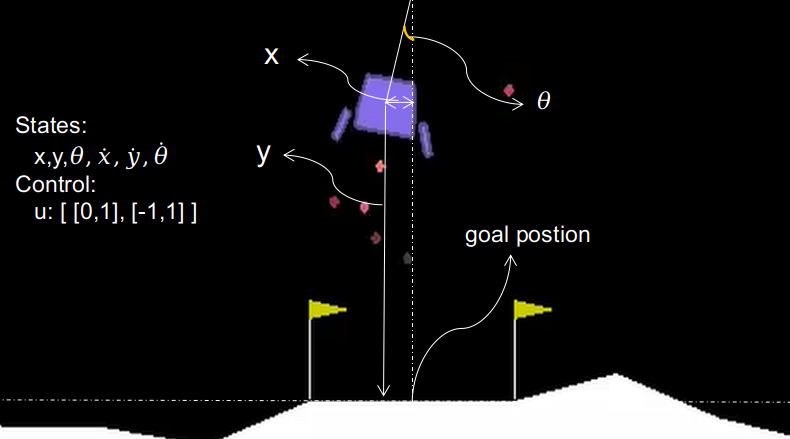}
    \vspace*{-4mm}
    \caption{Lunar Lander}\label{fig:lunar_lander_schematics}
    \vspace*{-4mm}
\end{figure}

As depicted in Figure \ref{fig:lunar_lander_schematics}, the lunar lander environment is a 6-dimensional state space with 2 control inputs. The state observables involve more terms and more closely related to a real-world game environment than previous classic dynamics examples.
\begin{equation}\label{eqn.state_space_ll}
    \begin{gathered}
        \boldsymbol{\chi}=[x,y,\theta, \dot{x}, \dot{y}, \dot{\theta} ] \\
        \boldsymbol{U}=[u_1, u_2], \quad {\rm where}, u_1\in[0,1], u_2\in[-1,1]
    \end{gathered}
\end{equation} 
where $x$, $y$ are the Cartesian coordinates of the lunar lander, and $\theta$ is the orientation angle measured clockwise from the upright position. $u_1$ represents the main engine on the lunar lander, which only takes positive input. In contrast, the $u_2$ is the side engine that can take both negative (left engine firing) and positive (right engine firing) values. The goal of this game is to move lunar lander from initial position ($x=10$, $y=13$) to the landing zone ($x=10$, $y=4$), subject to randomized initial gust disturbance and dynamics specified in the Box2D environment. 
The data collection can be obtained by running a un-trained reinforcement learning algorithm with random policy together with a random noise generated by the Ornstein–Uhlenbeck process during the data collection procedure. These randomization treatments are implemented to ensure enough exploration vs. exploitation ratio from the Reinforcement Learning model. In this study, we use 1876 data pairs obtained from playing five games to train our DKRC model.

\begin{figure}[h]
\centering
\includegraphics[width=0.35\textwidth]{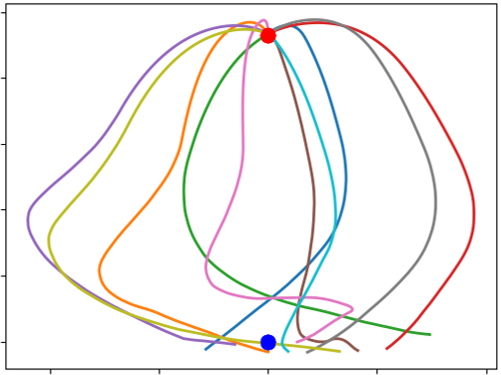}

\caption{Example 2D trajectories from 10 games played by DKRC. Red Point: Initial states, Blue Point: Goal states}\label{fig:lunarlander_traj}
\vspace{-6pt}
\end{figure}

The exact model for this system cannot be extracted directly from the OpenAI Gym environment but has to be identified either through a data-driven model-based approach (this paper) or model-free approach (e.g., reinforcement learning). A game score is calculated by OpenAI Gym where the system will reward smooth landing within the landing zone (area in between double flags) while penalizing the fuel consumption by engines ($u_1$, $u_2$).
The proposed method is first to generate the model of system identification. Assuming after the Koopman operator transform, the dynamical system can be considered to be linear. Therefore, model predictive control can be applied. The trajectories of the lunar lander successfully finishing the game are shown in Figure \ref{fig:lunarlander_traj}. Note that the trajectories for the ten games are spread out in the simulation environment and returning back to the same goal position. The reason for the spread-out behavior is that the initialization of each game will randommly assign an initial velocity and the control algorithm need to apply control to offset the drift while keeping balance in the air. In this simulation environment, the DKRC model was demonstrated to be able to learn the dynamics and cope with the unforeseen situation.
\vspace{-7pt}



\vspace{-7pt}


\section{Conclusions}
The proposed MPC controller designed by Deep Koopman Representation for Control (DKRC) has the benefit of being completely data-driven in learning, whereas it remains to be model-based in control design. The DKRC model is efficient in training and sufficient in the deployment in two OpenAI Gym environments compared to a state-of-the-art Reinforcement Learning algorithm (DDPG).

\bigskip

\bibliographystyle{IEEEtran}
\bibliography{refs,ref_Umesh}

\end{document}